\documentclass[margin=1in,letterpaper]{article}
\usepackage{amssymb}

\usepackage{url}
\usepackage{placeins}
\usepackage{colortbl} %
\usepackage{xparse}

\newlength{\fullimgw}
\newlength{\sliceimgw}
\AtBeginDocument{%
  \settowidth{\fullimgw}{\includegraphics{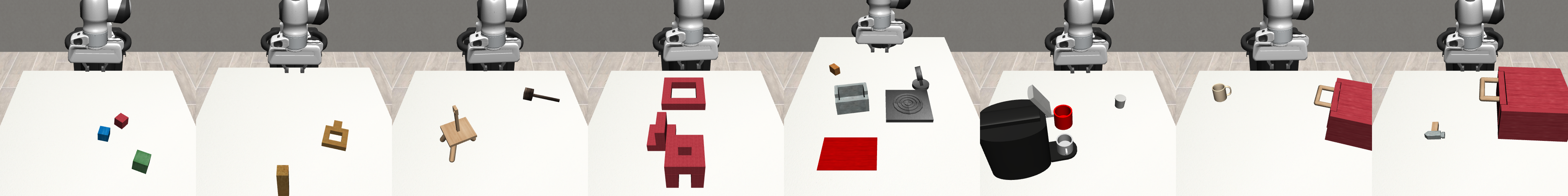}}%
  \setlength  {\sliceimgw}{\dimexpr \fullimgw/8\relax}%
}

\usepackage[
  textheight=9in,
  textwidth=5.5in,
  headheight=12pt,
  headsep=25pt,
  footskip=30pt
]{geometry}

\newcommand{\method}[1]{CP-Gen}
\usepackage{svg}

\usepackage{graphics} %
\usepackage{epsfig} %
\usepackage{mathptmx} %
\usepackage{times} %
\usepackage{amsmath} %
\usepackage{amssymb}  %
\usepackage{graphicx}
\usepackage{subcaption}
\usepackage{float}
\usepackage{xcolor}
\usepackage{color}
\usepackage{enumitem}

\usepackage{multirow}
\usepackage{booktabs}
\usepackage[font=footnotesize]{caption}
\usepackage{float}

\usepackage{algorithm}
\usepackage{algpseudocode}
\usepackage{amsfonts}
\usepackage{amssymb}
\usepackage{array}
\usepackage{tabularx}
\usepackage[hidelinks]{hyperref}

\usepackage[final]{corl_2025} %

\hypersetup{
    citecolor=orange,
}

\title{Constraint-Preserving Data Generation for Visuomotor Policy Learning}

\author{%
  \bfseries Kevin Lin$^{1,2}$ \quad Varun Ragunath$^{2*}$ \quad Andrew McAlinden$^{2*}$ \quad Aaditya Prasad$^1$ \\ \\
  \bfseries Jimmy Wu$^3$  \quad Yuke Zhu$^2$ \quad Jeannette Bohg$^1$\\ \\   
  $^1$Stanford University \quad $^2$University of Texas at Austin \quad $^3$Princeton University\\
  \footnotesize * Denotes equal contribution
}

\begin{document}
\maketitle

\vspace{-2em}
\begin{center}
  \textbf{\href{https://cp-gen.github.io}{cp-gen.github.io}}
\end{center}
\vspace{0.5em}

\begin{abstract}
Large-scale demonstration data has powered key breakthroughs in robot manipulation, but collecting that data remains costly and time-consuming. We present Constraint-Preserving Data Generation (CP-Gen), a method that uses a single expert trajectory to generate robot demonstrations containing novel object geometries and poses. 
These generated demonstrations are used to train closed-loop visuomotor policies that transfer zero-shot to the real world and generalize across variations in object geometries and poses. 
Similar to prior work using pose variations for data generation, CP-Gen first decomposes expert demonstrations into free-space motions and robot skills. But unlike those works, we achieve geometry-aware data generation by formulating robot skills as keypoint-trajectory constraints: keypoints on the robot or grasped object must track a reference trajectory defined relative to a task-relevant object.
To generate a new demonstration, \method{} samples pose and geometry transforms for each task-relevant object, then applies these transforms to the object and its associated keypoints or keypoint trajectories.
We optimize robot joint configurations so that the keypoints on the robot or grasped object track the transformed keypoint trajectory, and then motion plan a collision-free path to the first optimized joint configuration.
Experiments on 16 simulation tasks and four real-world tasks, featuring multi-stage, non-prehensile and tight-tolerance manipulation, show that policies trained using our method achieve an average success rate of 77\%, outperforming the best baseline which achieves an average of 50\%.

\end{abstract}

\section{Introduction}

Teaching robots to act by imitating humans is one of the most powerful, yet costly, ideas in robotics. At its best, large-scale imitation learning has enabled robots to tie shoelaces, fix broken grippers, and even cook shrimp~\cite{Zhao-RSS-23, fu2024mobile, zhao2024aloha, aldaco2024aloha}.
Further scaling promises to unlock broad generalization across complex manipulation tasks.
Behind these successes, however, lies a steep cost: human labor, months of continuous robot operation, and heavy infrastructure demands. 
For example, ALOHA Unleashed~\cite{zhao2024aloha} collected 26,000 demonstrations over 8 months across 10 robots and 35 operators, while DROID~\cite{khazatsky2024droid} gathered 76,000 real-world demonstrations over a year.
As the field pushes forward, reducing the burden of demonstration collection will be a key challenge.

Automated data generation promises to reduce the manual effort required in data collection.
One line of work \cite{mandlekar2023mimicgen, garrett2024skillgen, jiang2024dexmimicen, ameperosa2024rocoda, xue2025demogen} generates demonstrations for scenes with diverse object poses by collecting a handful of teleoperated demonstrations and algorithmically multiplying them using pose transformations and action replay.
Conceptually, these works aim to enable data efficient spatial generalization by exploiting $\mathrm{SE}(3)$ equivariance of robot actions with respect to object poses \cite{ameperosa2024rocoda}.
However, because they rely on $\mathrm{SE}(3)$ transformations, these works cannot generate demonstrations that adapt to geometric variations (such as different aspect ratios) with a given object instance.
For example, robot actions that succeed in hanging a short, wide wine glass onto a rack may fail entirely for a tall, narrow one, even if both are aligned under the same pose transformation.

A complementary strategy to ease the burden on manual data collection is to increase learning efficiency
by embedding structure---specifically, equivariance---into policy architectures ~\cite{wang22so2rl,wang21equivQ,huang2023leveraging,simeonov2023se,pan2023tax,huang2023edge,liu2023continual,jia2023seil,kim2023se,kohler2023symmetric,nguyen2023equivariant,nguyen2024symmetry,eisner2024deep}.
One line of work focuses on open-loop manipulation, exploiting symmetry to generalize across different object poses~\cite{rss22haojie,simeonov2021neural,ryu2023equivariant,huang2024fourier}.
More recent work has moved toward closed-loop policy learning, embedding equivariances directly into the policy architecture and directly predicting robot actions~\cite{rss22xupeng,wang22so2rl,zhu2023robot,wang2024equivariant,yang2024equibot,yang2024equivact}.
However, embedding equivariance bottlenecks can limit policy expressiveness and be insufficient when dealing with scenes that involve multiple interacting objects or symmetry-breaking variations.

In this work, we propose \textbf{Constraint-Preserving Data Generation (\textbf{CP-Gen}}), a geometry-aware data generation method that takes as few as a \emph{single} expert demonstration and generates new demonstrations with both novel object poses and novel geometries.
With \method{}, we aim to achieve the benefits of equivariant policy architectures without the bottleneck of embedding structure directly into policies.

\begin{figure*}[t]
  \centering
    \includegraphics[width=0.99\textwidth]{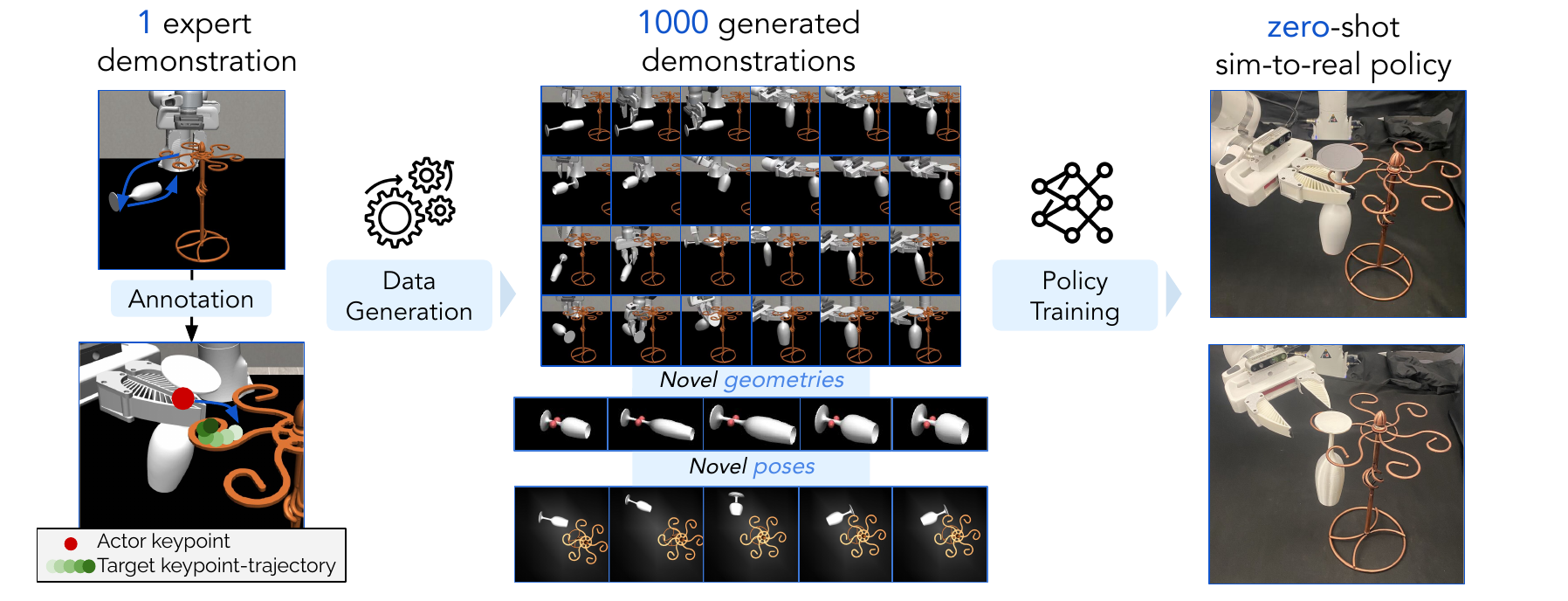}
  \captionof{figure}{\textbf{\method{}} uses one expert demonstration and keypoint-trajectory constraints to generate diverse demonstrations in simulation involving novel object geometries and poses, enabling large-scale policy training and zero-shot sim-to-real transfer. For the Wine Glass Spiral Hanging task, the first keypoint-trajectory is anchored to the wine glass and constrains the motion of points on the end effector. The second keypoint-trajectory is anchored relative to the spiral-shaped wine glass rack, and constrains a keypoint on the wine glass stem to track a complex spiral motion.
  }
    \vspace{-5mm}
  \label{fig:teaser}
\end{figure*}

Our key insight is to bring geometry awareness into the data generation process through a \textbf{keypoint-trajectory constraint} formulation.
Given an expert demonstration, we segment the trajectory into \emph{free-space motions} (segments replaceable with point-to-point collision-free motion planning) and \emph{robot skills} (segments requiring some interaction with task-relevant objects).
We then formulate each skill as a keypoint-trajectory constraint: selected robot or grasped-object keypoints must track a reference trajectory defined in the frame of a task-relevant object.
For example, in the \emph{Wine Glass Spiral Hanging} task (Figure~\ref{fig:teaser}), grasping is formulated as a trajectory-tracking constraint between keypoints on the gripper tips and a target pre-grasp trajectory in the wine-glass frame.  
The \emph{spiral} insertion introduces a second constraint: keypoints on the glass stem must track a spiral-like trajectory defined relative to the rack.
By anchoring keypoints to the reference frame of task-relevant objects, we can sample new object geometries and poses, adapt the keypoint-trajectory constraints according to the transforms applied to the original object geometry, and generate thousands of demonstrations covering diverse shapes and geometries. 
Using these demonstrations, we can train a visuomotor policy that naturally generalizes robustly across changes in object pose and geometry, without requiring additional human demonstrations.

Our contributions are threefold. First, we propose \method{}, a data generation method that produces robot demonstrations from a single expert demonstration using a keypoint-trajectory constraint formulation. Unlike prior work, and due to this formulation, \method{} generates data in a geometry aware manner.
Second, we validate our keypoint-trajectory data generation formulation on the MimicGen simulation benchmark. We achieve state-of-the-art performance (85\% vs. 63\%) on the default, pose-variation only task settings, as well as on our proposed simulation benchmark featuring novel object geometries (70\% vs. 37\%).
Third, we demonstrate successful zero-shot sim-to-real on a set of four real-world tasks featuring multi-stage, non-prehensile and tight-tolerance manipulation.

\section{Related Work}

\textbf{Data Generation for Robotics}.  
One approach to addressing data scarcity in robotics is to generate new demonstrations from teleoperated source data~\cite{mandlekar2023mimicgen,hoque2024intervengen,garrett2024skillgen,jiang2024dexmimicen, xue2025demogen}.  
MimicGen~\cite{mandlekar2023mimicgen} generates demonstrations by resetting object poses and replaying pose-transformed versions of the source trajectory.  
IntervenGen~\cite{hoque2024intervengen} extends MimicGen with human interventions mid-trajectory to correct failures, while DexMimicGen~\cite{jiang2024dexmimicen} adapts MimicGen for bimanual and dexterous settings.  
SkillGen~\cite{garrett2024skillgen} segments demonstrations into skills and motions, alternating between learned policies and motion planning to improve robustness.  
At the core of these methods is a relative-pose formulation and the exploitation of  $\mathrm{SE}(3)$-equivariance \cite{ameperosa2024rocoda}: given an  $\mathrm{SE}(3)$ transformation applied to an object, they apply the same  $\mathrm{SE}(3)$ transformation to robot actions to replicate the original effect.
As a result, they particularly struggle to generate demonstrations with new object geometries for tasks requiring tight-tolerance manipulation.  
\method{} addresses this limitation by introducing a keypoint-trajectory constraint formulation, representing skills in terms of robot or grasped-object keypoints tracking a reference trajectory.  
This formulation enables demonstration generation not only across novel object poses, but also across significantly different object geometries.

\textbf{Equivariance for Robotic Manipulation}.  
Another strategy to improve generalization is to embed geometric structure into policy inference and architecture.  
Open-loop methods design $\mathrm{SE}(2)$ or $\mathrm{SE}(3)$ equivariant feature representations on point clouds or images~\cite{wang22so2rl,wang21equivQ,huang2023leveraging,simeonov2023se,pan2023tax,huang2023edge,liu2023continual,jia2023seil,kim2023se,kohler2023symmetric,nguyen2023equivariant,nguyen2024symmetry,eisner2024deep}, achieving strong sample efficiency but often relying on expensive inference-time optimization~\cite{rss22haojie,simeonov2021neural,ryu2023equivariant,huang2024fourier} and lacking reactivity to dynamic changes.  
To address these limitations, recent work has moved toward closed-loop policies, beginning with planar $\mathrm{SO}(2)$ symmetries and extending to full $\mathrm{SIM}(3)$ equivariance~\cite{rss22xupeng,wang22so2rl,zhu2023robot,wang2024equivariant,yang2024equibot,yang2024equivact}, where $\mathrm{SIM}(3)$ captures rotation, translation, and uniform scaling transformations.  
While $\mathrm{SIM}(3)$-equivariant policies enable more flexible closed-loop control compared to $\mathrm{SE}(3)$-equivariant policies, they still fundamentally assume that object geometry variations can be captured by uniform scaling, and for larger geometry variations, they rely on policy generalization.
These closed-loop equivariance works \cite{wang2024equivariant, yang2024equivact, yang2024equibot} also tend to focus on single-object to robot equivariance, instead of the multi-object scenario.  
In \method{}, rather than using architectural equivariance, we leverage data generation to generalize to both novel poses and arbitrary geometry variations via a keypoint-trajectory constraint formulation.

\section{Problem Setting}

We are given a single expert demonstration in the form of a trajectory $\tau_{\text{src}} = \{o_t, a_t\}_{t=1}^H$ consisting of $H$ observations $o$, and $H$ actions $a$. Actions $a_t= \{a_{\text{eef}}, a_{\text{grip}}\}$ contain end-effector poses $a_{\text{eef}} \in \mathrm{SE}(3)$ along with gripper actions $a_{\text{grip}}$ for a given manipulation task. Observations $o_t = \{I_{\text{hand}}, I_{\text{3pv}}, q_t\}$ at each time step consist of a wrist-mounted camera image $I_{\text{hand}}$, a third-person view camera image $I_{\text{3pv}}$, and robot proprioception data $q_t$. We aim to train a visuomotor policy $\pi_\theta(a|o)$ mapping observations $o \in O$ to actions $a \in A$.

Similar to prior data generation methods using source demonstrations~\cite{mandlekar2023mimicgen, hoque2024intervengen, garrett2024skillgen}, we assume that an expert demonstration $\tau_{\text{src}}$ can be decomposed into a sequence of alternating free-space motion segments $\tau^{(i)}_{\text{motion}}$ and robot skill segments $\tau^{(i)}_{\text{skill}}$: $
\tau_{\text{src}} = [\tau^{(1)}_{\text{motion}}, \tau^{(1)}_{\text{skill}}, \tau^{(2)}_{\text{motion}}, \tau^{(2)}_{\text{skill}}, \ldots]$.
For each skill segment, we additionally assume access to a set of annotated keypoints on the robot gripper or a grasped object that are relevant to the completion of the skill. We refer to these as \emph{actor keypoints} and denote them by $\mathcal{K}_{\text{actor}}(t) = \{k^{\text{actor}}_1(t), \dots, k^{\text{actor}}_N(t)\}$, where each keypoint $k^{\text{actor}}_i(t) \in \mathbb{R}^3$ is defined in the local frame of the robot gripper or the grasped object at time $t$. Although in this work we manually selected these keypoints, in future work, keypoints can be predicted using vision-language models~\cite{huang2024rekep, oquab2023dinov2, openai2023gpt4}, or inferred via task-specific keypoint detectors~\cite{manuelli2019kpam, qin2020keto, sundaresan2020learning, manuelli2020keypoints}.

\begin{figure*}[t]
    \centering
    \includegraphics[width=\textwidth]{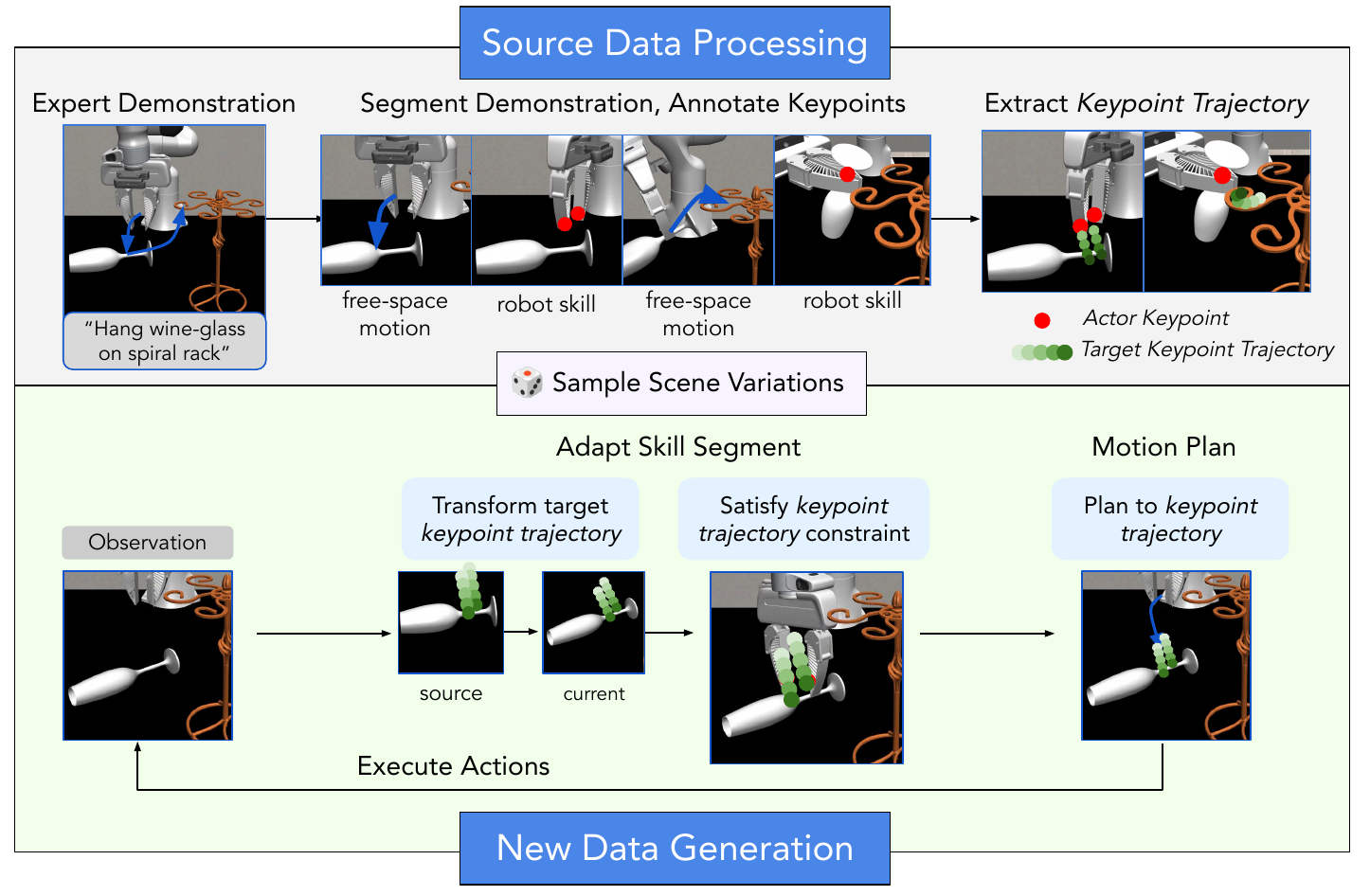}
    \caption{\textbf{\method{} Method}. 
    In the \emph{Source Data Processing} stage (top), starting from an expert demonstration $\tau_{\text{src}}$, we
    (a) segment the trajectory into free-space motion and skill segments,
    (b) annotate keypoints on the robot or grasped object \includegraphics[height=0.8em]{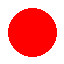}  (dubbed \emph{actor} keypoints), and
    (c) convert each skill segment into a \emph{keypoint-trajectory constraint} by extracting a keypoint-trajectory expressed in the frame of a task-relevant object.
    After processing the source demonstration, in the \emph{New Data Generation} stage (bottom), we 
    \includegraphics[height=0.8em]{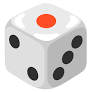}
    \textit{sample scene variations} by applying geometry and pose transforms to every task-relevant object.
    For each sampled scene, we
    (a) adapt our source skill segment to the current observation by transforming the extracted target keypoint-trajectory with the current object transforms and solving for robot configurations that satisfy the updated keypoint-trajectory constraint \includegraphics[height=0.8em]{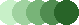}, and
    (b) motion plan a collision-free path that from the current robot joint configuration to the first configuration in the solved configuration trajectory. 
    The process iterates over all segments to generate a demonstration for the new scene, which can subsequently be used to train a visuomotor policy $\pi_\theta(a \mid o)$.}
    \vspace{-5mm}
    \label{fig:system}
\end{figure*}

\section{CP-Gen: Constraint-Preserving Data Generation}

We present \textbf{Constraint-Preserving Data Generation (\method{})}, a data generation method that enables one-shot visual imitation learning by adapting a source demonstration trajectory to scenes with novel object geometries and poses (see Figure~\ref{fig:system}).
Our approach begins by decomposing the original demonstration trajectory $\tau_{\text{src}}$ into a sequence of skill segments $\tau_{\text{skill}}$ interleaved with collision-free motion segments $\tau_{\text{motion}}$, such that $\tau_{\text{src}} = [\tau_{\text{motion}}, \tau_{\text{skill}}, \ldots, \tau_{\text{motion}}, \tau_{\text{skill}}]$. A skill segment is a segment that cannot be replaced by point-to-point collision-free motion planning, without affecting overall task success, while a motion segment is one that can be replaced by collision-free motion planning.
Each skill segment $\tau_{\text{skill}}$ is formulated as a \emph{keypoint-trajectory constraint}: keypoints on the robot or grasped object, \emph{actor keypoints} ${}^{O}\mathcal{K}_{\text{actor}}(t)$, must track a reference trajectory defined relative to a task-relevant object.
This reference trajectory is represented as a time-varying sequence of \emph{target keypoints} ${}^{O}\mathcal{K}_{\text{target}}(t)$ expressed in the local frame of the object $O$. We extract this reference keypoint-trajectory (Section~\ref{subsec:keypoint_extraction}) by transforming the actor keypoints into the object frame at each timestep. 
We assume that adapting and preserving this keypoint-trajectory constraint to a new scene leads to successful execution.
To generate a new demonstration in a novel scene, we first sample \textit{pose} and \textit{geometry} transforms for each task-relevant object.
We then apply the sampled transformations to the objects (and their associated keypoint trajectories) to produce a new scene reset.
For the new scene, we iterate through two phases of skill segment adaptation (Section~\ref{sec:skill-adaptation}) and motion planning (Section~\ref{sec:motion-planning}).
Finally, after filtering out any failed demonstrations using a success detector (which we assume exists for each task), we use imitation learning to train a visuomotor policy on the generated demonstrations.

\vspace{-2mm}
\subsection{Keypoint-Trajectory Constraint Extraction}
\label{subsec:keypoint_extraction}
\vspace{-2mm}

To extract the target keypoint-trajectory for a skill segment $\tau_{\text{skill}}$ from the original demonstration $\tau_{\text{src}}$, we transform the actor keypoints ${}^{A}\mathcal{K}_{\text{actor}} = \{{}^{A}k_1, \dots, {}^{A}k_N\}$, defined in the actor's local frame $A$ (robot gripper or grasped object), into the task-relevant object frame $O$ via a frame transform from actor to world frame, then world to task-relevant object frame:
\begin{equation}
{}^{O}k_i(t) = \underbrace{{}^{O}T_W(t)}_{\text{object pose}^{-1}} \cdot \underbrace{{}^{W}T_A(t)}_{\text{actor pose}} \cdot {}^{A}k_i,
\label{eq:keypoint_transform}
\end{equation}
Applying Eq.~\ref{eq:keypoint_transform} across all $t$ yields target keypoint-trajectory: ${}^{O}\mathcal{K}_{\text{target}}(t) = \{{}^{O}k_1(t), \dots, {}^{O}k_N(t)\}$.

\begin{figure*}[t]
  \centering

  \includegraphics[width=\textwidth]{figures/imgs/sim/tasks_D1.png}\\[1ex]
  {\scriptsize
    \makebox[0.125\textwidth][c]{StackThree}%
    \makebox[0.125\textwidth][c]{Square}%
    \makebox[0.125\textwidth][c]{Threading}%
    \makebox[0.125\textwidth][c]{Assembly}%
    \makebox[0.125\textwidth][c]{Kitchen}%
    \makebox[0.125\textwidth][c]{Coffee}%
    \makebox[0.125\textwidth][c]{MugCleanup}%
    \makebox[0.125\textwidth][c]{HammerCleanup}%
  }\\[2ex]

  \includegraphics[width=\textwidth]{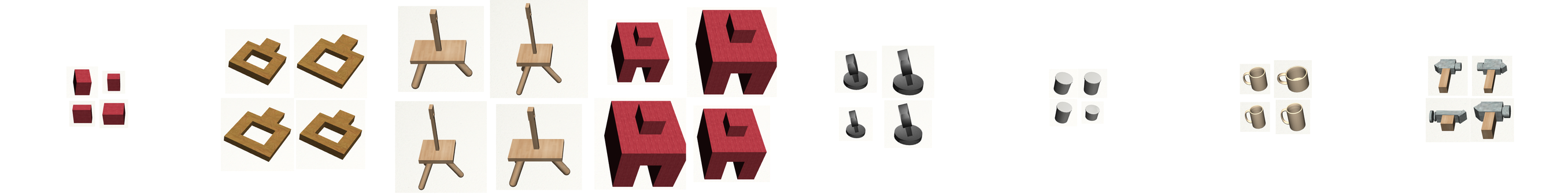}\\[1ex]
  {\scriptsize
    \makebox[0.125\textwidth][c]{StackThreeG}%
    \makebox[0.125\textwidth][c]{SquareG}%
    \makebox[0.125\textwidth][c]{ThreadingG}%
    \makebox[0.125\textwidth][c]{AssemblyG}%
    \makebox[0.125\textwidth][c]{KitchenG}%
    \makebox[0.125\textwidth][c]{CoffeeG}%
    \makebox[0.125\textwidth][c]{MugCleanupG}%
    \makebox[0.125\textwidth][c]{HmrCleanupG}%
  }\\[2ex]

\begin{minipage}{\textwidth}
  \centering
  \scriptsize
  \setlength{\tabcolsep}{6pt}
  \renewcommand{\arraystretch}{1.1}
  \begin{tabularx}{\textwidth}{l| *{8}{>{\centering\arraybackslash}X}}
    \toprule
    \textbf{Property} 
      & Stack Three & Square & Threading & Assembly 
      & Kitchen & Coffee & Mug Cleanup & Hammer Cleanup \\
    \midrule
    Object(s) 
      & block1, block2, block3 
      & nut, peg 
      & needle, tripod 
      & base, piece\_1, piece\_2 
      & bread, pot, stove, button 
      & pod, machine 
      & drawer, object 
      & hammer, drawer \\
    \midrule
    Scale Range 
      & [0.6,1.4] & [1.0,1.0] & [0.9,1.1] & [0.7,1.3] 
      & [0.7,1.4] & [0.6,1.1] & [0.7,1.3] & [0.7,1.2] \\
    \bottomrule
  \end{tabularx}
\end{minipage}

  \caption{\textbf{Simulation Tasks and Geometry Generalization Variants}. 
  \emph{Top}: Tasks from the MimicGen benchmark~\cite{mandlekar2023mimicgen} and object geometries sampled from our proposed \textit{Geometry Generalization} task variants. 
  \emph{Bottom}: Uniform scale sampling ranges applied in \textit{Geometry Generalization} task variants.}
  \label{fig:sim-tasks-and-scales}
  
  \vspace{-4mm}
\end{figure*}

\FloatBarrier

\vspace{-2mm}
\subsection{Skill Segment Adaptation}
\label{sec:skill-adaptation}
\vspace{-2mm}

Given a skill segment $\tau_{\text{skill}}$ and a current scene observation, our goal is to adapt the original keypoint-trajectory constraint to new task-relevant object geometries and poses. Doing so requires updating both the actor keypoints and the target keypoint-trajectory.

\textbf{Update Keypoint-Trajectory Constraint for New Geometry}.
Suppose that the task-relevant object undergoes a geometric transformation $\mathbf{X}$ (e.g., non-uniform scaling) in its own local frame $O$.
Let the original local-frame target keypoint-trajectory be ${}^{O}\mathcal{K}_{\text{target}}(t) = \{{}^{O}k_1(t), \dots, {}^{O}k_N(t)\}$. We therefore apply $\mathbf{X}$ to transform the target keypoint-trajectory:
\[
{}^{O}k'_i(t) = \mathbf{X} \cdot {}^{O}k_i(t), \quad \forall t.
\]

If the actor keypoints ${}^{A}\mathcal{K}_{\text{actor}}(t) = \{{}^{A}k_1(t), \dots, {}^{A}k_N(t)\}$ 
 are anchored to a grasped object that has been geometrically transformed, we similarly apply the object's local-frame geometric transformation $\mathbf{X}_{A}$ to the actor keypoints: ${}^{A}k'_i = \mathbf{X}_{A} \cdot {}^{A}k_i$.
These updated keypoints ${}^{A}k'_i$ are then used in the optimization process to match the newly transformed target trajectory.

\textbf{Solve for Robot Configuration via Keypoint Matching}.
Given the updated actor keypoints ${}^{A}k'_i$ (in the actor's local frame) and the updated target keypoint-trajectory ${}^{O}k'_i(t)$ (in the object’s local frame), our goal is to solve for the robot joint configuration $q_t^*$ at each timestep $t$ such that the transformed actor keypoints match the target keypoints in the world frame.

First, we compute the world-frame position of each actor keypoint (recall that actor keypoints are defined on the robot or a grasped object, and that we assume a fixed end-effector to grasped object frame transformation) via forward kinematics: ${}^{W}k'_i(q) = f_{\text{FK}}(q, {}^{A}k'_i)$.
Next, we compute the world-frame position of each target keypoint by transforming the updated local-frame keypoints using the current object pose ${}^{W}T_{O}(t)$ relative to the world frame: ${}^{W}k^{\text{target}}_i(t) = {}^{W}T_{O}(t) \cdot {}^{O}k'_i(t)$. 

Finally, we solve the following optimization problem for each timestep of the robot skill segment:
\begin{align}
q_t^* &= \arg\min_q
\underbrace{\sum_{i=1}^{N}\bigl\lVert f_{\mathrm{FK}}(q,\,{}^{A}k'_i)
- \bigl({}^{W}T_{O}(t)\cdot{}^{O}k'_i(t)\bigr)\bigr\rVert_2^2}_{\text{match keypoints}}
+ \underbrace{\lambda\,\bigl\lVert q - q^*_{t-1}\bigr\rVert_2^2}_{\text{temporal smoothness penalty}}
\label{eq:qp_objective}
\end{align}
where \(\lambda\ge0\) trades off keypoint matching and temporal smoothness in joint space.
This optimization encourages a robot trajectory where the transformed actor keypoints ${}^{A}k'_i$ track the updated keypoint-trajectory ${}^{O}k'_i(t)$ in the world frame under new object geometries and object poses.
We solve the optimization problem using the L-BFGS-B \cite{byrd95L-BFGS-B} gradient-based optimizer from SciPy \cite{2020SciPy-NMeth}.

\vspace{-2mm}
\subsection{Motion Planning}
\label{sec:motion-planning}
\vspace{-2mm}

After obtaining the optimized robot configurations $Q^* = [q_1^*, \ldots, q_H^*]$ from the skill segment adaptation described above, we plan a collision-free trajectory from the robot's current joint configuration to the first joint configuration of the upcoming skill segment $q_1^*$.
We query a collision-free motion planner \cite{curobo_report23, vamp_2024} with the robot's start configuration, the first joint configuration of the upcoming skill, and provide collision geometries extracted directly from our simulation environment.
We use an inverse-kinematics controller to track trajectories in our end-effector pose action space.

\vspace{-2mm}
\section{Experiments}
\vspace{-2mm}

Through our experiments, we wish to answer the following questions: 

\small 

\textbf{Q1}: Can a policy trained on data generated from a \emph{single demonstration} using \method{} generalize to \emph{unseen object geometries and poses}?

\textbf{Q2}: How does \method{}'s \emph{keypoint-trajectory formulation} affect \textit{data generation} success rates on tasks with \textit{novel object geometries}? 

\textbf{Q3}: How does training policies on \emph{diverse object geometries} affect performance in various settings?

\textbf{Q4}: Can \emph{pixel-based policies} trained with CP-Gen in simulation \emph{transfer zero-shot} to the real world?

\normalsize

\begin{table*}[t]
\centering
\scriptsize
\setlength{\tabcolsep}{4pt}
\begin{tabular}{l|cccccccc|c}
\toprule
\textbf{Method}   & StackThree    & Square         & Thread         & Assembly       & Kitchen        & Coffee         & Mug            & Hammer         & \textbf{Avg}  \\
\midrule
MimicGen [RGB]    & \textbf{0.98$\pm$0.01} & 0.72$\pm$0.04        & 0.38$\pm$0.04        & 0.32$\pm$0.04        & \textbf{0.98$\pm$0.01} & 0.58$\pm$0.04      & 0.78$\pm$0.03        & 0.64$\pm$0.04        & 0.67         \\
DemoGen [RGB]     & 0.79$\pm$0.03          & \textbf{0.87$\pm$0.03} & \textbf{0.87$\pm$0.03} & \textbf{0.85$\pm$0.03} & 0.88$\pm$0.03        & 0.98$\pm$0.01      & \textbf{0.89$\pm$0.03} & 0.79$\pm$0.03        & 0.87         \\
CP-Gen   [RGB]    & 0.82$\pm$0.03          & \textbf{0.87$\pm$0.03} & \textbf{0.87$\pm$0.03} & \textbf{0.85$\pm$0.03} & 0.96$\pm$0.01        & \textbf{1.00$\pm$0.00} & 0.86$\pm$0.03        & \textbf{0.80$\pm$0.03} & \textbf{0.88} \\
\midrule
MimicGen [D+S]    & 0.52$\pm$0.04          & 0.76$\pm$0.03        & 0.44$\pm$0.04        & 0.40$\pm$0.04        & \textbf{0.94$\pm$0.02} & 0.56$\pm$0.04      & 0.46$\pm$0.04        & 0.64$\pm$0.04        & 0.59         \\
DemoGen [D+S]     & \textbf{0.56$\pm$0.04} & 0.89$\pm$0.03        & \textbf{0.85$\pm$0.03} & \textbf{0.80$\pm$0.03} & 0.92$\pm$0.02        & \textbf{1.00$\pm$0.00} & \textbf{0.81$\pm$0.03} & 0.73$\pm$0.04        & \textbf{0.82} \\
CP-Gen   [D+S]    & \textbf{0.56$\pm$0.04} & \textbf{0.91$\pm$0.02} & \textbf{0.85$\pm$0.03} & 0.79$\pm$0.03        & 0.92$\pm$0.02        & 0.99$\pm$0.01      & 0.72$\pm$0.03        & \textbf{0.79$\pm$0.03} & \textbf{0.82} \\
\bottomrule
\end{tabular}
\vspace{2mm}
\vspace{2mm}
\begin{tabular}{l|cccccccc|c}
\toprule
\textbf{Method} & StackThreeG            & SquareG               & ThreadG              & AssemblyG                 & KitchenG             & CoffeeG             & MugG                & HammerG            & Avg \\
\midrule
MimicGen [RGB]  & 0.60$\pm$0.04          & 0.42$\pm$0.04         & 0.10$\pm$0.02        & 0.16$\pm$0.03            & 0.74$\pm$0.04        & 0.00$\pm$0.00       & 0.38$\pm$0.04       & 0.40$\pm$0.04       & 0.35 \\
DemoGen [RGB]   & 0.53$\pm$0.04          & 0.82$\pm$0.03         & 0.11$\pm$0.02        & 0.52$\pm$0.04            & 0.42$\pm$0.04        & 0.03$\pm$0.01       & 0.44$\pm$0.04       & 0.62$\pm$0.04       & 0.44 \\
CP-Gen [RGB]    & \textbf{0.68$\pm$0.04} & \textbf{0.88$\pm$0.03}& \textbf{0.79$\pm$0.03}& \textbf{0.55$\pm$0.04}   & \textbf{0.83$\pm$0.03}& \textbf{0.58$\pm$0.04}& \textbf{0.82$\pm$0.03}& \textbf{0.67$\pm$0.04}& \textbf{0.73} \\
\midrule
MimicGen [D+S]  & 0.22$\pm$0.03          & 0.64$\pm$0.04         & 0.05$\pm$0.02        & 0.38$\pm$0.04            & 0.84$\pm$0.03        & 0.00$\pm$0.00       & 0.50$\pm$0.04       & 0.45$\pm$0.04       & 0.39 \\
DemoGen [D+S]   & 0.17$\pm$0.03          & 0.74$\pm$0.04         & 0.11$\pm$0.02        & 0.52$\pm$0.04            & 0.32$\pm$0.04        & 0.04$\pm$0.02       & 0.42$\pm$0.04       & 0.59$\pm$0.04       & 0.36 \\
CP-Gen [D+S]    & \textbf{0.36$\pm$0.00} & \textbf{0.83$\pm$0.03}& \textbf{0.73$\pm$0.04}& \textbf{0.79$\pm$0.03}   & \textbf{0.86$\pm$0.03}& \textbf{0.63$\pm$0.04}& \textbf{0.76$\pm$0.04}& \textbf{0.74$\pm$0.04}& \textbf{0.67} \\
\bottomrule
\end{tabular}
\caption{\textbf{\method{} achieves state-of-the-art results on the MimicGen simulation benchmark}. \textbf{Top}:
On the original MimicGen benchmark (default \textit{Pose Only} task variants), \method{} and DemoGen perform similarly (average success rate of 85\% and 84.5\% respectively), compared to MimicGen's 63\% which uses neither a motion planner nor object-object data generation.
\textbf{Bottom}: On our custom benchmark containing \emph{Geometry Generalization} task variants which feature novel object geometries (denoted as TaskG), \method{} achieves an average success rate of 70\%, outperforming MimicGen and DemoGen by a margin of 33\% and 30\% respectively.
These results highlight \method{}'s strong generalization not only to pose variations but also to challenging geometric variations.
Bolded numbers indicate the best-performing method within each modality group. RGB denotes policies taking RGB image inputs; D+S denotes policies taking depth maps and segmentation mask inputs.
}
\vspace{-5mm}
\label{tab:sim-results}
\end{table*}

\subsection{Simulation Experiments}

\textbf{Tasks}. We evaluate all methods on single-arm Franka Panda tasks from the MimicGen~\cite{mandlekar2023mimicgen} benchmark.
We evaluate under two environment reset distributions. The default \textit{Pose Only} distribution varies only object poses, and corresponds to distribution ``D1" in the MimicGen benchmark.
The custom \textit{Geometry Generalization} distribution introduces shape variations via non-uniform scaling and retains the same pose reset range as the default \textit{Pose Only} distribution (Figure~\ref{fig:sim-tasks-and-scales}).

\textbf{Evaluation Metrics}. We report the average success rate achieved by a particular policy network on the given task, along with the standard error of this metric.
For computing standard errors, we use the best checkpoint rolled out for three seeds of 50 trials each.

\textbf{Baselines}. We compare \method{} to MimicGen \cite{mandlekar2023mimicgen} and DemoGen \cite{xue2025demogen}. MimicGen is a data generation method that produces demonstrations in which object poses differ from those seen in source demonstrations. 
 MimicGen first parses each source demonstration into object-centric subtask segments. 
 To generate a new demonstration, it selects a reference segment, transforms it to match the new scene’s object pose, runs linear interpolation to reach the start of the transformed reference segment and executes the resulting end-effector trajectory using a controller.
 Finally, MimicGen filters out failed runs using a custom success checker, in the same way as \method{}.

The authors additionally add Gaussian action noise to increase state coverage at the cost of the number of successfully generated demonstrations.
In our case, we use 0.01 noise by default, but set noise to 0 if the number of successfully generated demonstrations drops below 10\% for a given task.
We use MimicGen to generate 1000 successful demonstrations per task from 10 human demonstrations.

DemoGen generates synthetic demonstrations for novel object poses using a TAMP-based action generation method and 3D pointcloud editing. As the original setup does not leverage physics simulation, we adapt DemoGen to leverage a physics simulator for a fair comparison.
Like \method{}, DemoGen assumes a fixed grasp transform between the end-effector and object if the previous robot skill segment involves grasping an object.
DemoGen with physics can be viewed as \method{} without the keypoint trajectory constraints, and is implemented as such for our baseline.

\textbf{Policy Training}. For each method and task, we train a Diffusion Policy ~\cite{chi2023diffusionpolicy}, specifically the DDPM variant, on the corresponding generated dataset that contains 1000 successful demonstrations.
We use the same success classifier for \method{}, MimicGen and DemoGen to filter for successful demonstrations.
Policy input consists of end-effector pose, gripper width, and observations from a third-person and wrist-mounted camera. For camera observations, we use either RGB images or a depth map with segmentation masks.

\begin{figure*}[t]
    \begin{subfigure}[b]{0.24\textwidth}
        \centering
        \includegraphics[width=\textwidth]{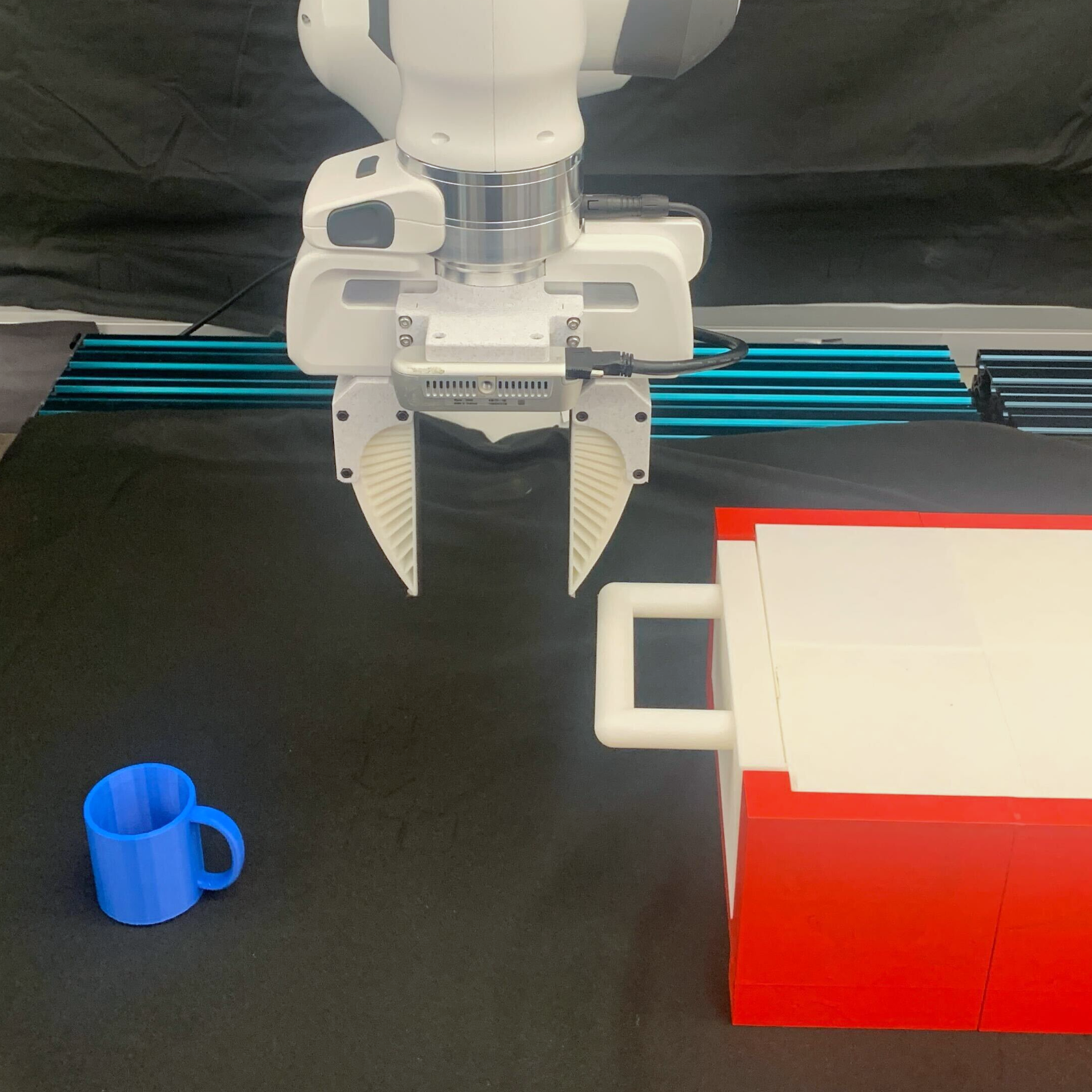}
        \caption{Mug Cleanup}
    \end{subfigure}
    \hfill
    \begin{subfigure}[b]{0.24\textwidth}
        \centering
        \includegraphics[width=\textwidth]{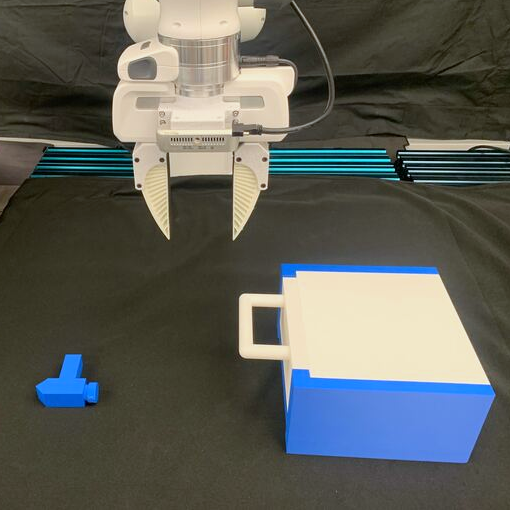}
        \caption{Hammer Cleanup}
    \end{subfigure}
    \hfill
    \begin{subfigure}[b]{0.24\textwidth}
        \centering
        \includegraphics[width=\textwidth]{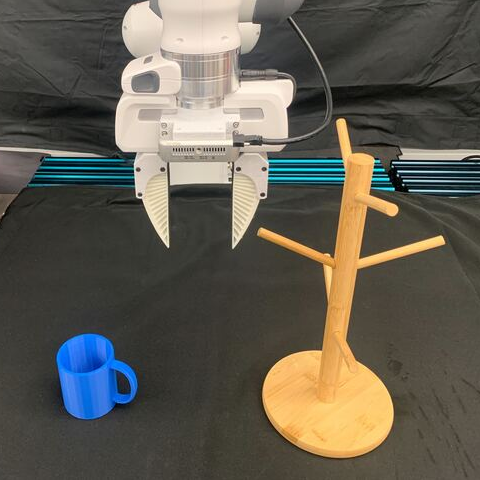}
        \caption{Mug Hanging}
    \end{subfigure}
    \hfill
    \begin{subfigure}[b]{0.24\textwidth}
        \centering
        \includegraphics[width=\textwidth]{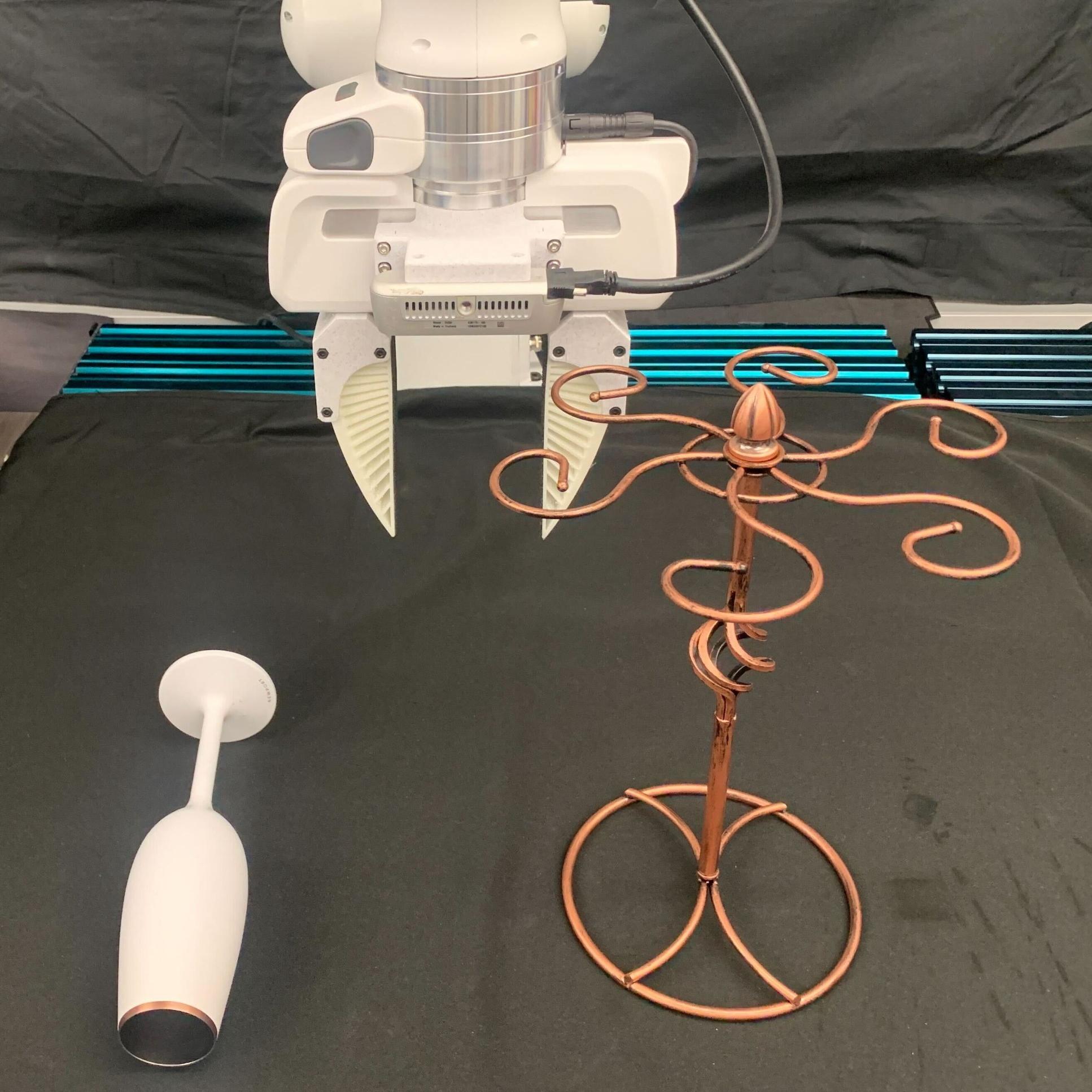}
        \caption{Wine Glass Spiral Hanging}
    \end{subfigure}
    \hfill
    \caption{\textbf{Real World Tasks}. We evaluate on four challenging real-world tasks and show that policies trained on  \method{} generated simulation datasets can transfer zero-shot to the real world.}
    \label{fig:real-tasks}
    \vspace{-8mm}
\end{figure*}

\begin{table*}[b]
\centering
\vspace{-5mm}
\scriptsize
\setlength{\tabcolsep}{8pt}
\renewcommand{\arraystretch}{1.1}
\begin{tabular}{l|cccc|c}
\toprule
\textbf{Method} & Mug Cleanup           & Hammer Cleanup        & Mug Hanging             & Wine Glass Spiral Hanging & \textbf{Average}          \\
\midrule
MimicGen & 0.20 $\pm$ 0.13           & 0.00 $\pm$ 0.00           & 0.80 $\pm$ 0.13           & 0.60 $\pm$ 0.16                      & 0.40    \\
CP-Gen   & \textbf{0.80 $\pm$ 0.13} & \textbf{0.80 $\pm$ 0.13} & \textbf{1.00 $\pm$ 0.00} & \textbf{0.70 $\pm$ 0.15} & \textbf{0.83}        \\
\bottomrule
\end{tabular}
\caption{\textbf{\method{} policies successfully transfer zero-shot sim2real}. Policies take in depth and segmentation masks, and tasks feature both object-geometry and object pose variations.}
\label{tab:real-results}
\end{table*}

\textbf{Results}. From Table \ref{tab:sim-results}, we see that policies trained with data generated by~\method{} (which takes a single source demonstration) outperform those trained using data generated using MimicGen and DemoGen across both observation input modalities.
On \textit{Pose Only} task variants, we see that \method{} outperforms policies trained using MimicGen data (85\% vs. 63\%).
We hypothesize  that the policy performance drop is due to dataset quality, and from two reasons.
First, \method{} uses a collision-free motion planner, and thus contains collision-avoidance behaviors, while MimicGen data does not have explicit collision-avoidance behaviors.
Second, \method{} uses \textit{actor keypoints} (keypoints on the robot or grasped object) during data generation which enables the robot to adapt actions based on how an object is grasped, while MimicGen only preserves relative $\mathrm{SE}(3)$ transforms between the robot and a reference object.
For example, on the Threading task, regardless of how the robot grasps the needle, MimicGen would generate the same end-effector actions in the threading phase, based on the tripod's pose. In contrast, \method{} would adjust the robot's actions to align the actor keypoints on the needle tip with the tripod. In the \textit{Pose Only} setting, CP-Gen is equivalent to our implementation of DemoGen and thus achieves similar success rates.
On the \textit{Geometry Generalization} task variants, \method{}'s advantage is more pronounced (70\% vs. 37\% and 40\%).
This finding is unsurprising, as \method{} generated demonstrations adapted to changes in object geometries, while MimicGen and DemoGen do not consider such geometry changes during data generation.

\vspace{-2mm}
\subsection{Real World Experiments}
\vspace{-2mm}

\textbf{Tasks}. We evaluate \method{} and baselines on four real-world manipulation tasks (Figure~\ref{fig:real-tasks}):
(1) \textit{Mug Cleanup}: Open a drawer, pick and place mug  into a drawer, and close the drawer.
(2) \textit{Hammer Cleanup}: Same as \textit{Mug Cleanup}, except with a small hammer, introducing grasping challenges.
Cleanup tasks test multi-stage, non-prehensile manipulation involving articulated objects.
(3) \textit{Mug Hanging}: Pick and insert a mug onto a branch-like hook. This task tests constrained insertion through a small opening.
(4) \textit{Wine Glass Spiral Hanging}: Pick, reorient and hang a wine glass by threading the stem through a spiral-shaped rack.
This task tests tight tolerance path following and orientation control.
For each task, we manually construct a digital twin in simulation, generate 1000 successful demonstrations using a single source demonstration, and train a Diffusion Policy on the dataset. Policies are evaluated on two novel object geometries for each task.

\textbf{Policy Inputs}. We use end-effector poses, gripper width and two camera observations. Specifically, we use depth and segmentation masks for the third-person-view camera, and segmentation masks for the hand camera. We use depth and segmentation masks to reduce the sim-to-real gap \cite{dalal2024manipgen}.

\textbf{Results}. From Table~\ref{tab:real-results}, we see that CP-Gen achieves strong zero-shot sim-to-real transfer on all four manipulation tasks. In contrast, MimicGen struggles with novel geometries, as it generates fewer successful demos there. In the \textit{Cleanup} tasks, failures arise from misalignment during drawer opening. In the \textit{Hanging} tasks, failures are caused by suboptimal grasp positioning, which prevents proper insertion.
Task failures did not come from collisions with non-relevant objects, so we posit that the performance gain came from the geometry-aware demonstration data generated by \method{}.

\begin{table*}[t]
\centering
\scriptsize
\vspace{-5mm}
\begin{tabular}{
  >{\centering\arraybackslash}p{0.45\textwidth}|
  >{\centering\arraybackslash}p{0.45\textwidth}
}
\toprule
\multicolumn{1}{c|}{\textbf{(a) Data Generation}} &
\multicolumn{1}{c}{\textbf{(b) Policy Evaluation}} \\
\midrule
\begin{tabular}{lcc}
  \textbf{Method}                & \textbf{Pose} & \textbf{Geometry} \\
  \midrule
  MimicGen                       & 0.58                & 0.23                     \\
  CP-Gen                         & 0.89                & 0.62                     \\
\end{tabular}
&
\begin{tabular}{lcc}
  \textbf{Method}                & \textbf{Pose} & \textbf{Geometry} \\
  \midrule
  CP-Gen (Pose)            & 0.88                & 0.45                     \\
  CP-Gen (Geometry)       & 0.72                & 0.73                     \\
\end{tabular}
\\
\bottomrule
\end{tabular}
\caption{\textbf{Ablation Results.}
(a) CP-Gen outperforms MimicGen in data generation under both \textit{pose}-only and \textit{geometry} generalization settings.
(b) Generating varied geometries (\textit{geometry}) boosts policy generalization to novel objects (73\% vs.\ 45\%).}
\label{tab:ablation-results}
\vspace{-6mm}
\end{table*}

\FloatBarrier

\vspace{-2mm}
\subsection{Ablations}
\label{sec:ablations}
\vspace{-2mm}

We conduct a series of ablations to study two aspects of \method{}: the use of the keypoint-trajectory constraint formulation, and the effect of training on diverse object geometries.
First, to assess the impact of the keypoint-trajectory constraint formulation, we compare data generation success rates between \method{} and MimicGen (which uses a relative pose data generation formulation) across both \textit{Pose Only} and \textit{Geometry Generalization} task variants on eight simulation tasks.
For each task, we run 50 data generation trials.
In Table~\ref{tab:ablation-results}(a), CP-Gen achieves consistently higher success rates (89\% vs. 58\%, 62\% vs. 23\%), especially on the \textit{Geometry Generalization} reset distribution, highlighting the value of the geometry-aware data generation that our keypoint-trajectory constraint formulation enables. Note that our comparison with DemoGen \cite{xue2025demogen} in Table \ref{tab:sim-results} also serves as an ablation for using keypoint trajectory constraints versus only using pose transforms, as that is the difference between CP-Gen and our DemoGen baseline. Keypoint-trajectory constraints lead to a 30\% increase in policy success rates for CP-Gen vs. DemoGen in the Geometry Generalization setting.
Second, we evaluate how training on diverse object geometries affects policy generalization. 
In Table~\ref{tab:ablation-results}(b), we report policy success rates when training with and without geometry diversity, and evaluated on the \textit{Pose} and \textit{Geometry Generalization} environment reset settings.
As expected, training with diverse object geometries substantially improves policy generalization to novel geometries (73\% vs. 45\%).
We also observe that training and evaluating on the same distribution yields the best evaluation performance (73\% vs. 45\% and 88\% vs. 72\%).

\vspace{-2mm}
\section{Conclusion} 
\label{sec:conclusion}
\vspace{-2mm}

In this work, we propose \method{}, a data generation framework for generalizable visual imitation learning given just one expert demonstration.
\method{} generates geometry-aware demonstrations using the insight that robot skills can be formulated as keypoint-trajectory constraints: keypoints on the robot or grasped
object must track a reference trajectory defined relative to a task-relevant object.
We demonstrate that \method{}'s approach to data generation from just one source demonstration achieves state-of-the-art success rates on the MimicGen simulation benchmark, and outperforms baselines on a custom benchmark featuring diverse object geometries.
Finally, we show successful zero-shot sim-to-real transfer of policies trained with \method{} data to challenging real-world tasks.

\clearpage

\textbf{Limitations}. CP-Gen requires manual annotation of robot skill segments and keypoints.
The simulation environment and reward functions are also manually defined; future work could incorporate foundation model based success classifiers.
The method assumes a fixed skill sequence, limiting task level generalization.
Incorporating task level planning via foundation models \cite{saycan-2022, innermono-2022, lin2023text2motion, liu2023llm+} or task and motion planner methods \cite{kaelbling2011integrated, garrett2020-pddlstream, integrated-tamp-2021} can further improve data generation success.
Current results are limited to a single-arm Panda robot, though the method can be extended to multi-arm, mobile base, and legged systems \cite{jiang2024dexmimicen}. \method{} does not handle the situation where one may want demonstrations generated for a specific object instance within the same category for which a source demonstration is available. For example, we may have a source demonstration for a mug with a square handle, but wish to generate demonstrations for a specific mug that might have a round handle.
\method{} would need require the geometric transformation from the square handle mug to the round handle mug, which may be non-trivial to obtain. 
Future work may explore the use of neural descriptor fields \cite{simeonov2021neural} to assist in this setting.

\acknowledgments{Toyota Research Institute provided funds to support this work. Additionally, this work was partially supported by the National Science Foundation (FRR-2145283, EFRI-2318065), the Office of Naval Research (N00014-24-1-2550), the DARPA TIAMAT program (HR0011-24-9-0428), and the Army Research Lab (W911NF-25-1- 0065). It was also supported by the Institute of Information \& Communications Technology Planning \& Evaluation (IITP) grant funded by the Korean Government (MSIT) (No. RS-2024-00457882, National AI Research Lab Project). Finally, we thank Timothy Chen, Wil Thomason, Zak Kingston, Tyler Lum, Priya Sundaresan, Fanyun Sun, Yifeng Zhu, Zhenyu Jiang, Mingyo Seo, Megan Hu, William Chong, Marion Lepert, and Brent Yi for helpful discussions throughout the project.}

\bibliography{example.bib}       

\end{document}